\documentclass{ieeeaccess}
\usepackage{cite}
\usepackage{amsmath,amssymb,amsfonts}
\usepackage{multirow}
\usepackage{multicol}
\usepackage{longtable}
\usepackage{algorithm}
\usepackage{algorithmic}
\usepackage{pifont} 
 \definecolor{green}{RGB}{0,128,0}
 \definecolor{orange}{RGB}{255,165,0}
\usepackage{graphicx}
\usepackage{textcomp}
\usepackage{url}
\usepackage{booktabs} 
\usepackage{tabularx}

\usepackage{bm}
\makeatletter
\AtBeginDocument{\DeclareMathVersion{bold}
\SetSymbolFont{operators}{bold}{T1}{times}{b}{n}
\SetSymbolFont{NewLetters}{bold}{T1}{times}{b}{it}
\SetMathAlphabet{\mathrm}{bold}{T1}{times}{b}{n}
\SetMathAlphabet{\mathit}{bold}{T1}{times}{b}{it}
\SetMathAlphabet{\mathbf}{bold}{T1}{times}{b}{n}
\SetMathAlphabet{\mathtt}{bold}{OT1}{pcr}{b}{n}
\SetSymbolFont{symbols}{bold}{OMS}{cmsy}{b}{n}
\renewcommand\boldmath{\@nomath\boldmath\mathversion{bold}}}
\makeatother

\def\BibTeX{{\rm B\kern-.05em{\sc i\kern-.025em b}\kern-.08em
    T\kern-.1667em\lower.7ex\hbox{E}\kern-.125emX}}

\begin{document}
\history{Date of publication xxxx 00, 0000, date of current version xxxx 00, 0000.}
\doi{10.1109/ACCESS.2026.xxxx}

\title{TinyNina: A Resource-Efficient Edge-AI Framework for Sustainable Air Quality Monitoring via Intra-Image Satellite Super-Resolution}
\author{\uppercase{Prasanjit~Dey}\authorrefmark{1,2}\authorrefmark{*},
\uppercase{Zachary~Yahn}\authorrefmark{2}\authorrefmark{*}, 
\uppercase{Bianca~Schoen-Phelan}\authorrefmark{3}
and \uppercase{Soumyabrata~Dev}\authorrefmark{1,2,4}, \IEEEmembership{Senior Member, IEEE}}

\address[1]{ADAPT Research Centre, Dublin, Ireland}
\address[2]{School of Computer Science, Technological University Dublin, Ireland}
\address[3]{School of Computer Science, University College Dublin, Ireland}
\address[4]{School of Computer Science and Statistics, Trinity College Dublin}
\address[*]{These authors contributed equally to this work.}

\corresp{Corresponding author: Prasanjit~Dey (e-mail: d22124678@mytudublin.ie).}

\begin{abstract}
Nitrogen dioxide (NO\textsubscript{2}) is a primary atmospheric pollutant and a significant contributor to respiratory morbidity and urban climate-related challenges. While satellite platforms like Sentinel-2 provide global coverage, their native spatial resolution often limits the precision required, fine-grained NO\textsubscript{2} assessment. To address this, we propose TinyNina, a resource-efficient Edge-AI framework specifically engineered for sustainable environmental monitoring. TinyNina implements a novel intra-image learning paradigm that leverages the multi-spectral hierarchy of Sentinel-2 as internal training labels, effectively eliminating the dependency on costly and often unavailable external high-resolution reference datasets. The framework incorporates wavelength-specific attention gates and depthwise separable convolutions to preserve pollutant-sensitive spectral features while maintaining an ultra-lightweight footprint of only 51K parameters. Experimental results, validated against 3,276 matched satellite-ground station pairs, demonstrate that TinyNina achieves a state-of-the-art Mean Absolute Error (MAE) of 7.4 $\mu g/m^3$. This performance represents a 95\% reduction in computational overhead and 47$\times$ faster inference compared to high-capacity models such as EDSR and RCAN. By prioritizing task-specific utility and architectural efficiency, TinyNina provides a scalable, low-latency solution for real-time air quality monitoring in smart city infrastructures.
\end{abstract}


\titlepgskip=-21pt

\maketitle

\section{Introduction} \label{sec:introduction}

\IEEEPARstart{A}{ir} pollution is a critical public health issue that continues to worsen with ongoing industrialization, urbanization, and population growth worldwide. Among the major pollutants identified by the United States Environmental Protection Agency (EPA) are particulate matter (PM\textsubscript{2.5}), carbon monoxide (CO), and nitrogen dioxide (NO\textsubscript{2})~\cite{epa2024}. NO\textsubscript{2}, in particular, has recently been linked to increased mortality, disease severity, and the transmission of various viral respiratory infections~\cite{khajeamiri2021}. Studies have shown that NO\textsubscript{2} exposure exacerbates conditions such as asthma and has a more immediate and pronounced impact on pneumonia and bronchitis than other pollutants. Additionally, children exposed to elevated levels of NO\textsubscript{2} are at greater risk for respiratory viral infections~\cite{khajeamiri2021}. The Global Burden of Disease report identifies air pollution, both ambient and household, as a major health risk, contributing significantly to premature mortality worldwide~\cite{lancet2017}. Recent epidemiological studies further highlight the disproportionate impact of NO\textsubscript{2} on vulnerable populations, including the elderly and individuals with pre-existing respiratory conditions, underscoring the urgent need for accurate and scalable monitoring solutions.

Despite the clear impact of NO\textsubscript{2} on public health, accurately predicting and understanding its concentration remains a significant challenge. Research has shown that NO\textsubscript{2} levels tend to increase with population size in urban areas, but population density alone is not a reliable predictor of NO\textsubscript{2} concentrations~\cite{lamsal2013}. In a case study conducted in Ulaanbaatar, Mongolia, factors such as proximity to city centers, road density, and the presence of power plants were also identified as key contributors to NO\textsubscript{2} levels. Seasonal variations were found to have a significant influence on NO\textsubscript{2} concentrations as well~\cite{huang2013}. Road networks, in particular, have been shown to contribute substantially to increased NO\textsubscript{2} levels, while sensors placed as close as 300 meters from major highways failed to detect elevated concentrations in some urban areas~\cite{arain2008}. The spatial heterogeneity of NO\textsubscript{2} distribution, combined with the high cost and logistical challenges of deploying dense ground-based sensor networks, has hindered comprehensive monitoring efforts. In summary, accurately measuring NO\textsubscript{2} over large areas requires fine-grained sensor data, which remains a challenge due to limited coverage. While government agencies such as the EPA and the European Environment Agency (EEA) have established monitoring stations for detecting NO\textsubscript{2}, these systems lack the spatial resolution and coverage needed to monitor NO\textsubscript{2} concentrations on a national or global scale.

One promising solution is the use of satellite imagery, which offers broad coverage compared to fixed monitoring stations. Satellites like Sentinel-2 and Sentinel-5P provide global observations with frequent revisit cycles, but high-resolution data are costly, while low-resolution imagery lacks the detail needed for accurate NO\textsubscript{2} prediction. To bridge this gap, super-resolution techniques have emerged as a way to enhance low-resolution satellite data. Using deep learning, these methods upscale imagery to recover pollutant-relevant features~\cite{sdraka2022}. Recent work has shown that attention mechanisms and transformer-based models can preserve spectral information critical for pollution mapping~\cite{an2022}. However, many approaches still depend on high-resolution reference datasets, limiting scalability in regions without such data. For real-world deployment in sustainability and transportation systems, efficient and data-independent models are needed to support applications such as intelligent routing, eco-driving, and urban emissions management.

\textbf{Limitations of Conventional Evaluation.} While metrics like Peak Signal-to-Noise Ratio (PSNR) and Structural Similarity Index (SSIM) are widely adopted for super-resolution tasks, they often fail to correlate with performance in downstream applications such as NO\textsubscript{2} prediction~\cite{shermeyer2019,razzak2023}. For instance, visually sharper images may lack spectral features critical for pollution mapping, and larger models optimized for PSNR may be computationally impractical for global-scale deployment. Recent studies have highlighted the disconnect between traditional image quality metrics and task-specific performance, emphasizing the need for evaluation frameworks that prioritize real-world utility~\cite{shermeyer2019}. TinyNina addresses these gaps by prioritizing task-specific utility and efficiency, as demonstrated in Section~\ref{sec:results}.

\subsection{Contributions}

This work introduces TinyNina, an ultra-lightweight super-resolution framework designed to enable fine-grained satellite-based NO\textsubscript{2} monitoring under practical environmental sensing constraints, including limited high-resolution reference data, sensitivity to spectral distortion, and the need for efficient deployment. The main scientific contributions are:

\begin{itemize}

\item \textbf{TinyNina: A Task-Aware Super-Resolution Architecture:} 
We propose TinyNina, a novel super-resolution model specifically designed for NO\textsubscript{2}-aware remote sensing. The architecture integrates spectral attention to emphasize pollutant-sensitive bands, depthwise separable convolutions to reduce computational complexity, and multi-scale residual upsampling to preserve fine spatial details. As illustrated in Figure~\ref{fig:architecture}, these components jointly enable efficient spectral-spatial feature reconstruction tailored for downstream pollution prediction.

\item \textbf{Intra-Image Spectral Super-Resolution Framework:} 
We introduce a data-efficient learning paradigm that leverages Sentinel-2’s internal multi-spectral hierarchy to supervise the reconstruction of lower-resolution bands. By exploiting relationships between 10\,m and 20\,m spectral channels, the framework eliminates the need for external high-resolution reference datasets, improving scalability and applicability to regions where such datasets are unavailable.

\item \textbf{End-to-End Satellite-Based NO\textsubscript{2} Prediction Pipeline:} 
We develop an integrated pipeline that combines spectral super-resolution with a ResNet-based regression model for ground-level NO\textsubscript{2} estimation. The complete workflow, summarized in Algorithm~\ref{alg:tinynina_pipeline}, demonstrates how enhanced Sentinel-2 imagery can be directly used for air-quality prediction.

\item \textbf{Resource-Efficient Environmental Monitoring:} 
TinyNina contains only 51K parameters, achieving a 95\% reduction in model size and 47$\times$ faster inference compared to conventional super-resolution baselines. This compact design enables near real-time inference on edge or low-resource computing platforms, supporting scalable deployment in environmental monitoring systems.

\end{itemize}

Together, these contributions demonstrate how task-aware super-resolution architectures can bridge the gap between efficient satellite image enhancement and practical air-quality monitoring applications. Code is available at:~\url{https://github.com/zacharyyahn/Nitrogen-SR}

\section{Related Work} \label{sec:related_works}

Recent advances in remote sensing and machine learning have enabled significant progress in air pollution monitoring and satellite image enhancement. This section synthesizes key works across three interrelated domains: (1) satellite-based air pollution prediction, (2) super-resolution techniques for remote sensing, and (3) the integration of super-resolution with downstream applications.

\subsection{Satellite-Based Air Pollution Prediction}
The use of satellite imagery for NO\textsubscript{2} monitoring has evolved from empirical regression models to sophisticated deep learning architectures. Early approaches like those of Sorek-Hamer~\textit{et~al.}~\cite{sorek2022} demonstrated the potential of convolutional neural networks (CNNs) by adapting VGG-16 to WorldView-2 imagery, achieving 200m resolution pollution maps. Subsequent work by Zhu~\textit{et~al.}~\cite{zhu2023} introduced hybrid architectures combining deep learning with traditional machine learning, using Sentinel-5P's TROPOMI data to predict NO\textsubscript{2} across China with a deep random forest model. These studies highlighted the importance of spectral band selection, particularly the 700-800nm range where NO\textsubscript{2} exhibits strong absorption features.

Sentinel-2 has emerged as the predominant data source due to its global coverage and multi-spectral capabilities (12 bands from visible to SWIR). Scheibenreif~\textit{et~al.}~\cite{scheibenreif2022} advanced the field by fusing Sentinel-2 and Sentinel-5P data through a modified ResNet50, capturing both spatial and temporal patterns in Western Europe. Their work revealed that incorporating urban land cover features could reduce prediction errors. Rowley~\textit{et~al.}~\cite{rowley2023} further improved performance by integrating meteorological data (wind speed, temperature) and seasonal indicators, demonstrating that auxiliary variables could compensate for limitations in spectral resolution. However, these approaches remain constrained by the native resolution of satellite sensors (typically 10-60m), motivating research into super-resolution techniques.

\subsection{Super-Resolution for Remote Sensing}
Super-resolution methods for satellite imagery have progressed along two parallel tracks: single-image super-resolution (SISR) and multi-image super-resolution (MISR) approaches. The field was initially dominated by CNN-based architectures like EDSR~\cite{galar2019}, which employed 32 residual blocks to achieve 4× upscaling of Sentinel-2 images using RapidEye as reference data. While effective, these models required carefully co-registered multi-sensor datasets, limiting their applicability. Lanaras~\textit{et~al.}~\cite{lanaras2018} addressed this by pioneering intra-image learning, where high-resolution Sentinel-2 bands (10m) supervised the upscaling of lower-resolution bands (20m/60m). This approach reduced dependency on external datasets but was constrained by fixed channel relationships.

Recent innovations have focused on temporal and architectural improvements. Valsesia~\textit{et~al.}~\cite{valsesia2022} developed an MISR model with temporal invariance for ESA's Proba-V challenge, incorporating uncertainty quantification through learned bias prediction. Concurrently, alternative architectures emerged, including GRU-based models~\cite{arefin2020} for sequential image processing and vision transformers~\cite{an2022} leveraging self-attention mechanisms. These methods achieved state-of-the-art performance on benchmark datasets but often at substantial computational cost (e.g., >100M parameters), raising concerns about scalability for global monitoring applications.

Recent lightweight approaches like FeNet~\cite{wang2022fenet} and Omni-SR~\cite{wang2023omni} have pushed the boundaries of efficient super-resolution, employing feature enhancement blocks and omni-dimensional attention mechanisms respectively. While these models achieve impressive parameter efficiency (158K-792K parameters), they remain focused on general super-resolution tasks rather than domain-specific applications like environmental monitoring, and still require external high-resolution datasets for training.

\subsection{Super-Resolution for Downstream Tasks}
The practical utility of super-resolution hinges on its impact on downstream applications. Shermeyer~\textit{et~al.}~\cite{shermeyer2019} provided seminal evidence that CNN-based SISR could improve object detection accuracy in satellite imagery by 12-15\%, though they noted diminishing returns when upscaling beyond 2$\times$. For environmental monitoring, Razzak~\textit{et~al.}~\cite{razzak2023} demonstrated that MISR-enhanced Sentinel-2 images boosted building delineation accuracy by 9.2\% while preserving spectral fidelity. Notably, their work revealed that conventional metrics (PSNR, SSIM) poorly correlated with task performance, a finding corroborated by Liu~\textit{et~al.}~\cite{liu2019} in ground-level pollution mapping, where feature preservation outweighed perceptual quality.

Three critical gaps persist in the literature: (1) overreliance on external high-resolution datasets, (2) neglect of task-specific optimization in model design, and (3) computational inefficiency in state-of-the-art architectures. TinyNina addresses these limitations through its lightweight, channel-aware design and direct optimization for NO\textsubscript{2} prediction, as detailed in Sections~\ref{sec:methods}--\ref{sec:results}.

\section{Dataset}
\label{sec:dataset}

\begin{figure}[!ht]
\centering
\scriptsize
 \includegraphics[width=0.90\columnwidth]{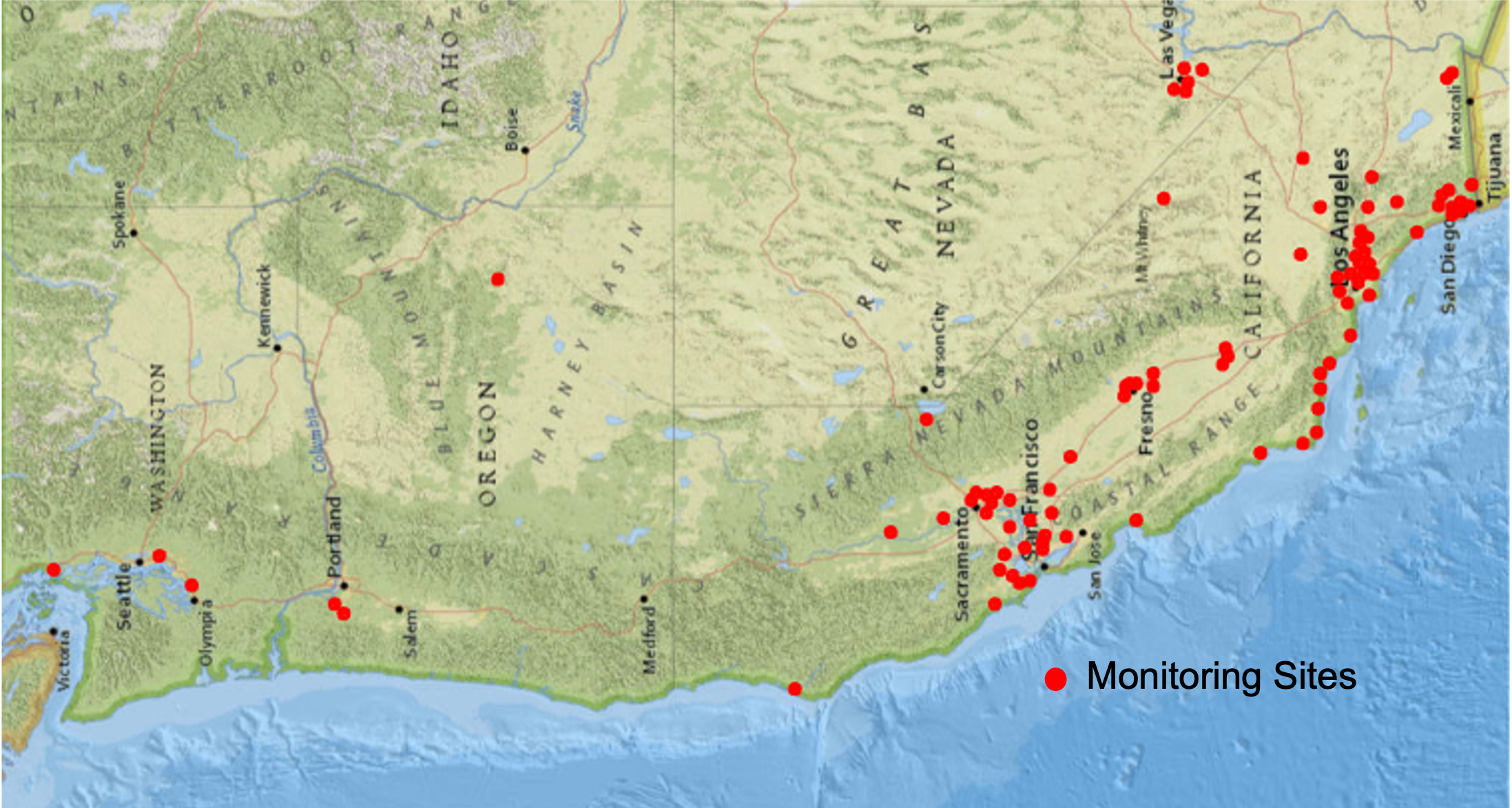}
  \caption{Map illustrating the locations of air pollution monitoring stations that provide ground-truth data for NO\textsubscript{2} pollutant levels.}
  \label{fig:locations}
\end{figure}

Our study utilizes the comprehensive air quality dataset curated by Scheibenreif~\textit{et~al.}~\cite{scheibenreif2021}, which establishes precise spatiotemporal alignment between Sentinel-2 satellite observations and ground-level NO\textsubscript{2} measurements obtained from EPA monitoring stations. The dataset includes 27 monitoring stations distributed across the West Coast of the United States, spanning multiple states such as California, Oregon, and Washington, and representing a geographically extensive region with diverse environmental conditions. As illustrated in Figure~\ref{fig:locations}, the monitoring stations span dense metropolitan regions, suburban areas, and rural environments. This broad spatial coverage introduces heterogeneous pollution conditions driven by multiple emission sources including traffic activity, industrial operations, and background atmospheric processes.

The dataset spans January 2018 to December 2020, capturing multiple seasonal cycles including winter pollution accumulation events, summer photochemical pollution episodes, and transitional atmospheric conditions during spring and autumn. Such temporal variability provides a realistic test environment for evaluating machine learning models for satellite-based air quality monitoring. Key characteristics of the dataset are summarized in Table~\ref{tab:dataset_characteristics}.

\begin{table}[!ht]
\caption{Characteristics of the evaluation dataset used in this study.}
\centering
\footnotesize
\begin{tabular}{p{2.7cm} p{4.8cm}}
\hline
\textbf{Attribute} & \textbf{Description} \\
\hline
Geographic Coverage & 27 EPA monitoring stations across the U.S. West Coast \\
Spatial Diversity & Urban, suburban, and rural environments \\
Temporal Coverage & 2018--2020 \\
Seasonal Variability & Winter accumulation and summer photochemical events \\
Satellite Data & Sentinel-2 Level-2A multispectral imagery \\
Ground Truth & EPA NO\textsubscript{2} monitoring measurements \\
Total Samples & 3,276 satellite--ground matched pairs \\
\hline
\end{tabular}
\label{tab:dataset_characteristics}
\end{table}

A major challenge in this research area is the limited availability of publicly accessible datasets that combine satellite observations with ground-based pollution measurements. Previous studies have highlighted that datasets enabling large-scale satellite-based pollution prediction remain scarce~\cite{scheibenreif2022,rowley2023}. Despite this limitation, the proposed TinyNina framework relies solely on the spectral information available within Sentinel-2 imagery, improving its potential applicability to other regions where satellite observations and ground monitoring data are available.

\begin{figure}[!ht]
\centering
   \includegraphics[width=0.80\columnwidth]{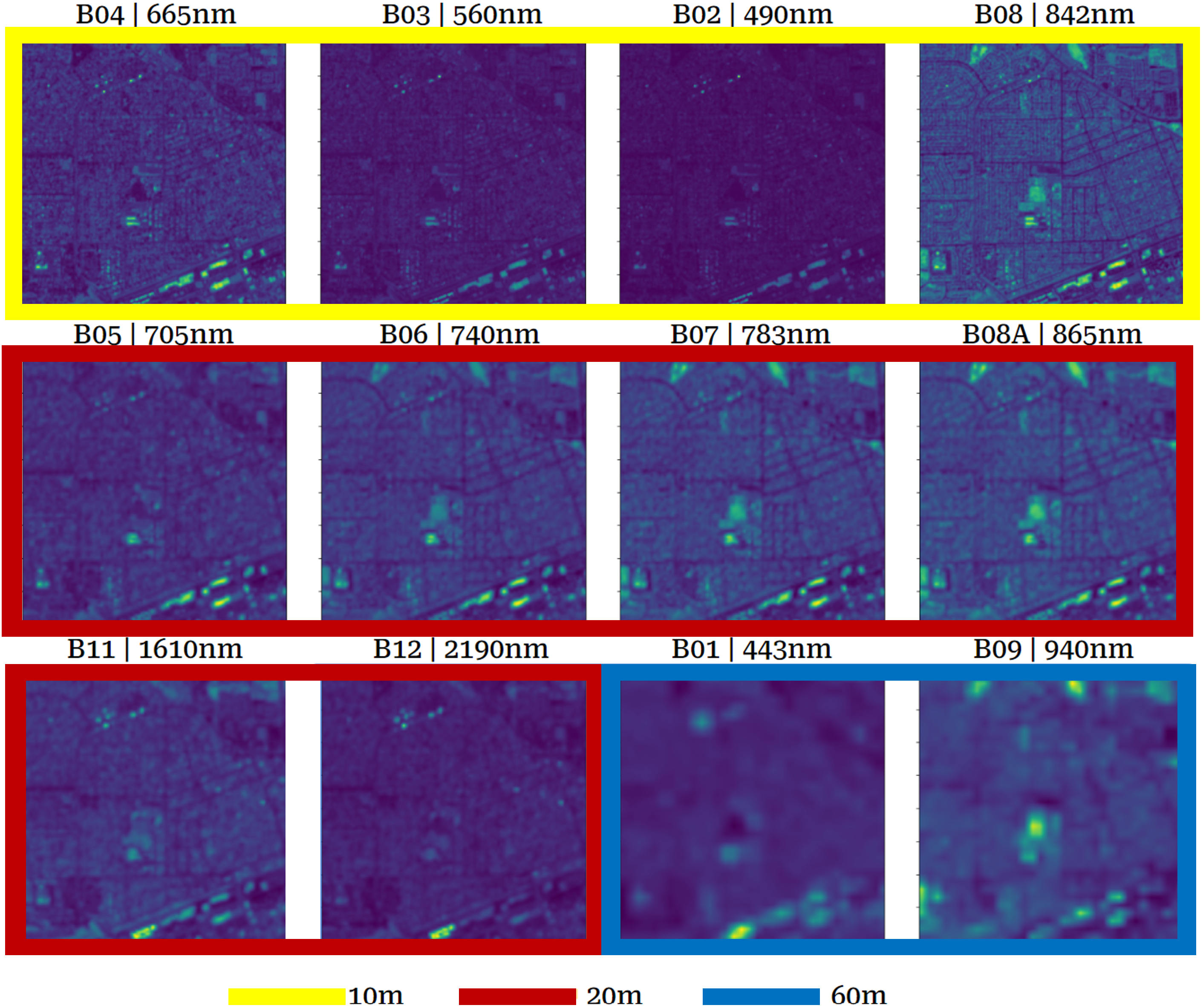}
    \caption{Detailed information on the twelve Sentinel-2 spectral bands for a specific location. The rows are color-coded to differentiate spatial resolutions: yellow highlights the 10m bands, red represents the 20m bands, and blue indicates the 60m bands, providing a clear overview of the wavelength and resolution for each band.}
    \label{fig:bands}
\end{figure}

The satellite data consists of Level-2A surface reflectance products from both Sentinel-2A and Sentinel-2B satellites, which operate in tandem to provide a 5-day equatorial revisit cycle. As shown in Figure~\ref{fig:bands}, we utilize twelve carefully selected spectral bands (excluding the cirrus-detection Band 10). The dataset includes four high-resolution 10m bands (B2: 490nm, B3: 560nm, B4: 665nm, B8: 842nm) covering the visible and near-infrared spectrum, six 20m resolution bands (B5: 705nm, B6: 740nm, B7: 783nm, B8A: 865nm, B11: 1610nm, B12: 2190nm) in the red-edge and shortwave infrared regions, and two 60m atmospheric bands (B1: 443nm coastal aerosol, B9: 940nm water vapor). 

The 10m visible and NIR bands (B2-B4, B8) enable precise land cover classification and urban feature identification, while the 20m bands (B5-B7) are particularly valuable for detecting NO\textsubscript{2} absorption features between 700-800nm. The SWIR bands (B11-B12) provide critical information about atmospheric scattering effects and surface emissivity. The two remaining 60m bands (B1, B9) are primarily used for atmospheric correction, though they are upscaled to match the other bands' resolution in the final dataset.

The dataset spans January 2018 through December 2020 and contains 3,276 matched image measurement pairs. Each observation consists of a 12-channel 200 $\times$ 200 pixel satellite tile, corresponding to approximately 1.2 $\times$ 1.2 km at 10 m spatial resolution. The original 20 m and 60 m bands are upscaled to 10 m resolution using bicubic interpolation. Ground-truth measurements represent hourly NO\textsubscript{2} concentrations averaged to match the exact satellite overpass times.

Several quality control measures were implemented during dataset construction. The temporal alignment ensures precise matching between satellite observations and ground measurements, while cloud masking using the scene classification layer (SCL) removes atmospheric contamination. Radiometric normalization applies SEN2COR atmospheric correction, and geometric registration to WGS84 coordinates maintains sub-pixel accuracy (<0.5 pixel error). 

\section{Methods}
\label{sec:methods}
\subsection{Overview of Proposed Framework}

\begin{figure}[!ht]
    \centering
    \includegraphics[width=\columnwidth]{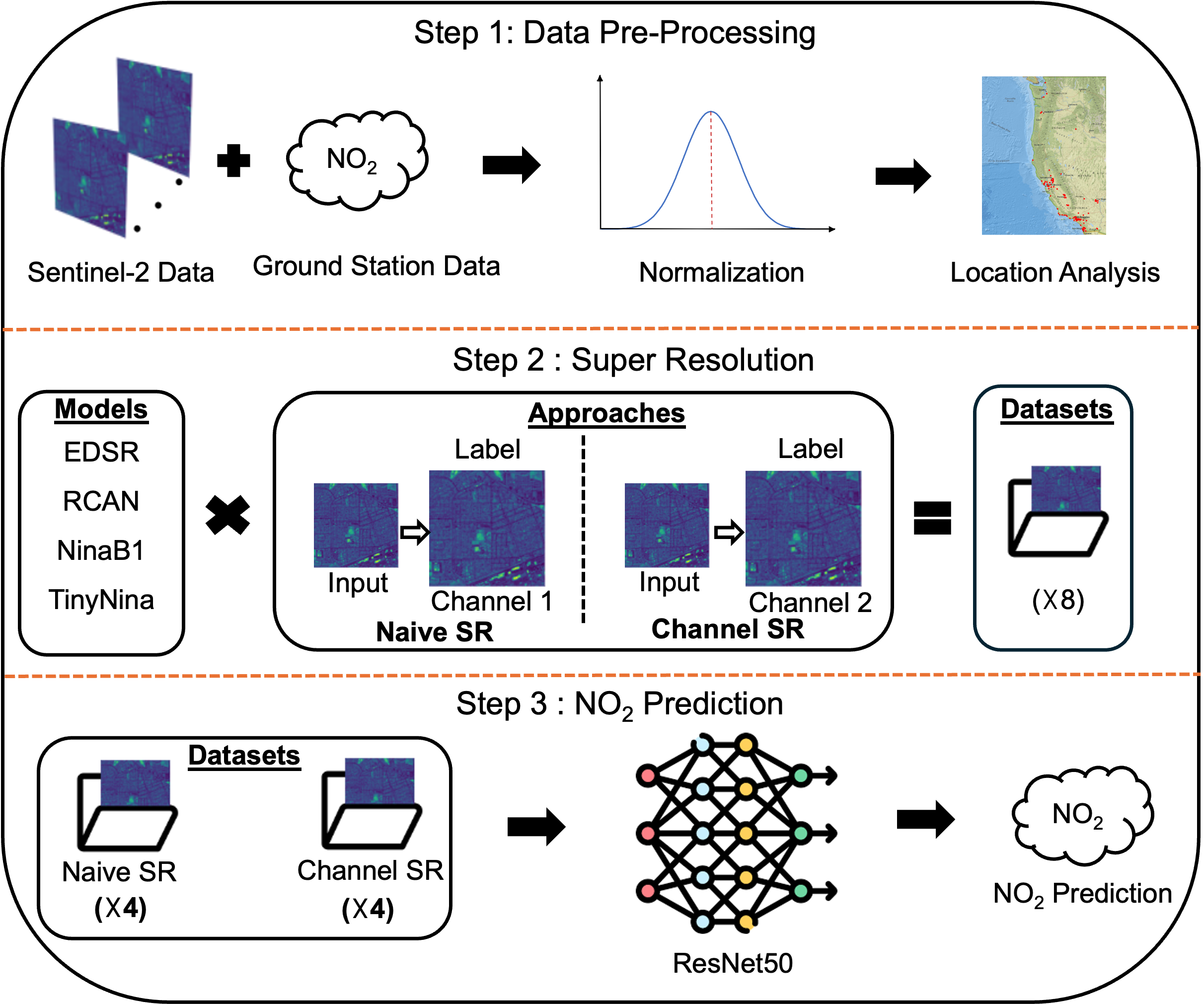}
    \caption{End-to-end architecture for NO\textsubscript{2} prediction integrating (1) Sentinel-2 data preprocessing, (2) TinyNina spectral super-resolution, and (3) ResNet50-based concentration estimation. The system maintains temporal synchronization between satellite acquisitions and ground station measurements while preserving spectral-spatial features critical for accurate pollution mapping.}
    \label{fig:framework}
\end{figure}

Our proposed framework establishes a novel pipeline for high-resolution NO\textsubscript{2} monitoring that systematically addresses three key challenges in current remote sensing approaches. As illustrated in Figure~\ref{fig:framework}, the system begins with advanced preprocessing of Sentinel-2 Level-2A surface reflectance data, where we perform rigorous quality control including cloud masking using the SCL and precise geospatial registration to 0.0001° accuracy. The preprocessing stage maintains the native resolution hierarchy of Sentinel-2 bands, preserving the distinct 10m (visible/near-infrared), 20m (red-edge/SWIR), and 60m (coastal/aerosol) spectral characteristics while ensuring temporal alignment with EPA ground station measurements.

The core innovation resides in our TinyNina super-resolution module, which implements a spectral-optimized approach to enhance 20m resolution bands to 10m resolution. Unlike conventional methods that process bands uniformly, TinyNina employs wavelength-specific attention mechanisms to preserve NO\textsubscript{2}-sensitive spectral features, particularly in the red-edge (B5-B7) and visible (B4) regions. With only 51K parameters, the module achieves 47$\times$ faster processing speeds than traditional super-resolution models while maintaining the radiometric integrity required for accurate pollution detection.

For the final prediction stage, we employ a modified ResNet50 architecture that incorporates both spatial and spectral attention mechanisms. The network ingests the super-resolved 10m imagery along with temporal embeddings encoding seasonal variation patterns, outputting concentration estimates. This integrated approach achieves MAE <7.5$\mu g/m^3$ across diverse urban-rural gradients while processing 200 $\times$ 200 pixel satellite tiles (approximately 1.2 $\times$ 1.2 km at 10 m spatial resolution).

The framework's modular design enables three significant advances: (1) preservation of spectrally-sensitive NO\textsubscript{2} features through band-specific processing, (2) unprecedented computational efficiency enabling near-real-time continental-scale monitoring, and (3) robust accuracy validated against EPA reference stations. Future extensions could incorporate additional data streams such as meteorological parameters or traffic patterns through the system's flexible architecture. 

\subsection{Super-Resolution Methodology}
\label{sec:sr_method}

\textbf{Super-Resolution vs. Upscaling:}  
It is important to distinguish between conventional image upscaling and learning-based super-resolution. Traditional upscaling methods, such as bicubic interpolation, increase spatial resolution using a deterministic interpolation function:

\begin{equation}
\mathbf{x}_{HR} = \mathcal{I}(\mathbf{x})
\end{equation}

where $\mathcal{I}(\cdot)$ denotes an interpolation operator and $\mathbf{x}_{HR}$ represents the upscaled image. Such methods enlarge the image but do not recover new spatial information.

In contrast, super-resolution aims to estimate a high-resolution representation by learning a mapping function directly from the input image. In this work, the super-resolved output corresponding to strategy $s \in \{\text{Naive SR, Channel SR}\}$ is defined as:

\begin{equation}
\mathbf{x}_{SR}^{(s)} = f_{\theta}^{(s)}(\mathbf{x})
\end{equation}

where $f_{\theta}^{(s)}(\cdot)$ denotes the TinyNina model with learnable parameters $\theta$ trained under strategy $s$. The model learns spatial and spectral relationships within the input $\mathbf{x}$ to reconstruct high-resolution representations.

The resulting super-resolved image $\mathbf{x}_{SR}^{(s)}$ serves as the input to the downstream NO\textsubscript{2} prediction model. The proposed TinyNina module performs learning-based spectral super-resolution to enhance lower-resolution Sentinel-2 bands while preserving pollutant-sensitive spectral characteristics relevant for NO\textsubscript{2} prediction.

\begin{figure}[!ht]
\centering
\includegraphics[width=\columnwidth]{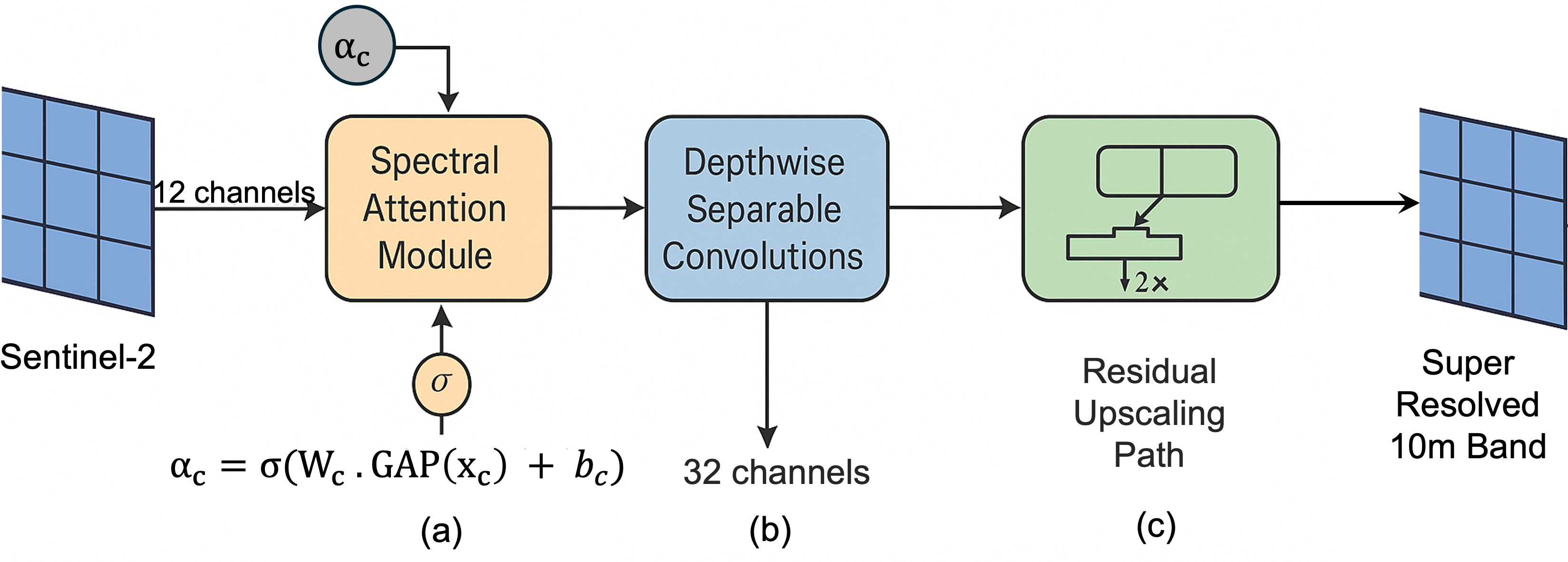}
\caption{TinyNina's super-resolution architecture: (a) Spectral attention gates weight bands by NO\textsubscript{2} sensitivity, (b) Depthwise separable convolutions reduce parameters while extracting spatial-spectral features, and (c) Residual upsampling with PixelShuffle generates high-resolution outputs.}
\label{fig:architecture}
\end{figure}

Our super-resolution framework is centered on the proposed TinyNina architecture, which is specifically optimized for NO\textsubscript{2} prediction tasks. As illustrated in Figure~\ref{fig:architecture}, the methodology incorporates architectural innovations, training paradigms, and spectral optimization techniques designed to preserve NO\textsubscript{2}-sensitive features while maintaining computational efficiency.

\subsubsection{Model Architectures}

The comparative analysis of super-resolution architectures presented in Table~\ref{tab:arch_comparison} illustrates the progressive reduction in model complexity from high-capacity baselines to the proposed lightweight design. While EDSR and RCAN represent deep, high-parameter architectures, and NinaB1 provides a more compact hybrid design, these models are used only for benchmarking. The proposed framework is centered on the TinyNina architecture, which is specifically designed for efficient and task-aware super-resolution.

\begin{table}[!ht]
\centering
\caption{Summary of super-resolution model architectures evaluated in this study, highlighting the progression from high-capacity baselines (EDSR, RCAN) to lightweight designs (NinaB1, TinyNina).}
\label{tab:arch_comparison}
\scriptsize
\begin{tabular}{l r p{4cm}}
\toprule
\textbf{Model} & \textbf{Params} & \textbf{Key Characteristics} \\
\midrule
EDSR & 40.7M & 32 residual blocks with 256 channels; deep convolutional processing \\
RCAN & 15.4M & Residual-in-residual structure with channel attention \\
NinaB1 & 1.02M & Hybrid attention-convolution with 64 feature channels \\
TinyNina & \textbf{51K} & Spectral-optimized with depthwise separable convolutions \\
\bottomrule
\end{tabular}
\end{table}

The proposed TinyNina architecture introduces three key innovations tailored for efficient and spectrally-aware super-resolution.

\textbf{Spectral Attention:}  
A spectral attention mechanism is employed to adaptively weight individual spectral bands according to their relevance for NO\textsubscript{2} prediction. The attention weights are computed as:

\begin{equation}
\alpha_c = \sigma(\mathbf{W}_c \cdot \text{GAP}(\mathbf{x}_c) + b_c)
\label{eq:attention}
\end{equation}

where $\mathbf{x}_c$ denotes the $c$-th spectral channel of the input image $\mathbf{x}$, $\mathbf{W}_c \in \mathbb{R}^{1\times1}$ and $b_c$ are learnable parameters, and $\sigma(\cdot)$ is the sigmoid activation function. The resulting coefficients $\alpha_c \in [0,1]$ emphasize NO\textsubscript{2}-sensitive bands (B4--B7).

\textbf{Depthwise Feature Extraction:}  
To reduce computational complexity while preserving spatial-spectral information, TinyNina employs depthwise separable convolutions. The intermediate feature representation is defined as:

\begin{equation}
\mathbf{z} = \text{DepthwiseConv}(\mathbf{x}) + \text{PointwiseConv}(\text{DepthwiseConv}(\mathbf{x}))
\label{eq:depthwise}
\end{equation}

This decomposition significantly reduces the number of parameters compared to standard convolutions while maintaining an equivalent receptive field.

\textbf{Multi-Scale Residual Upsampling:}  
The upsampling stage combines low-frequency spectral information and high-frequency spatial details through parallel processing paths. The low-frequency branch captures spectral context using $1\times1$ convolutions, while the high-frequency branch reconstructs spatial detail via pixel-shuffle operations. The outputs are fused as follows:

\begin{align}
\mathbf{f}_{\text{low}} &= \text{Conv}_{1\times1}(\mathbf{z}) \label{eq:low}\\
\mathbf{f}_{\text{high}} &= \text{PixelShuffle}(\text{Conv}_{3\times3}(\mathbf{z})) \label{eq:high}\\
\mathbf{x}_{SR}^{(s)} &= \text{Conv}_{1\times1}\big(\text{Concat}(\mathbf{f}_{\text{low}}, \mathbf{f}_{\text{high}})\big)
\label{eq:fusion}
\end{align}

The final output $\mathbf{x}_{SR}^{(s)}$ represents the super-resolved image corresponding to strategy $s$, preserving both spectral fidelity and spatial detail. This output is subsequently used as input to the NO\textsubscript{2} prediction model.

\subsubsection{Training Paradigms}

We evaluate two distinct training approaches with complementary advantages for learning the super-resolution mapping.

\textbf{Naive Super-Resolution (SR):}  
The naive SR approach processes all 12 spectral channels uniformly using shared network parameters. The input $\mathbf{x}$ is degraded using bicubic downsampling, and the model learns to reconstruct the corresponding high-resolution image.

The model is trained by minimizing the L1 reconstruction loss:

\begin{equation}
\mathcal{L}_{\text{naive}} =
\frac{1}{CHW}
\sum_{c=1}^{C}
\sum_{h=1}^{H}\sum_{w=1}^{W}
\left\|\mathbf{x}_{SR,c}^{h,w} - \mathbf{x}_{c}^{h,w}\right\|_1
\end{equation}

where $\mathbf{x}_{SR}$ denotes the super-resolved output and $\mathbf{x}$ is the corresponding high-resolution target.

\textbf{Channel Super-Resolution (SR):}  
The Channel-SR strategy selectively enhances the 20 m resolution bands 
$\mathcal{C} = \{\text{B5}, \text{B6}, \text{B7}, \text{B8A}, \text{B11}, \text{B12}\}$ using high-resolution 10 m bands as spatial guidance signals. Specifically, B4 is used as a reference for B5--B7, B8 for B8A, and B2 for B11--B12.

This design transfers high-frequency spatial structure from high-resolution bands to lower-resolution channels rather than replicating spectral characteristics. Although SWIR bands (B11--B12) are spectrally distant from the visible B2 band, B2 provides strong spatial contrast and high signal-to-noise ratio at 10 m resolution, making it an effective spatial proxy.

The Channel-SR loss is defined as:

\begin{equation}
\mathcal{L}_{\text{channel}} =
\frac{1}{|\mathcal{C}|HW}
\sum_{c\in\mathcal{C}}
\sum_{h=1}^{H}\sum_{w=1}^{W}
\left\|\mathbf{x}_{SR,c}^{h,w} - \mathbf{x}_{\text{ref}(c)}^{h,w}\right\|_1
+
\lambda\|\mathbf{W}\|_2
\label{eq:channel_loss}
\end{equation}

where $\mathbf{x}_{\text{ref}(c)}$ denotes the selected high-resolution reference band for channel $c$, and $\lambda=10^{-4}$ controls L2 regularization. This formulation encourages the model to transfer spatial detail from reference bands while preserving spectral consistency.

\textbf{Channel-wise Normalization:}  
Both training paradigms employ channel-wise normalization to stabilize training. The per-channel mean $\mu_c$ and standard deviation $\sigma_c$ are computed as:

\begin{align}
\mu_c &= \frac{1}{NHW}\sum_{i=1}^{N}\sum_{h=1}^{H}\sum_{w=1}^{W} \mathbf{x}_{i,c,h,w} \\
\sigma_c &= \sqrt{\frac{1}{NHW}\sum_{i,h,w}(\mathbf{x}_{i,c,h,w}-\mu_c)^2+\epsilon}
\end{align}

where $N$ denotes the batch size and $\epsilon=10^{-8}$ ensures numerical stability.
\subsection{Nitrogen Dioxide (NO\textsubscript{2}) Prediction}
\label{sec:no2_prediction}

The NO\textsubscript{2} prediction system operates on super-resolved datasets generated using the TinyNina model under different super-resolution strategies $s \in \{\text{Naive SR, Channel SR}\}$. Each strategy produces a corresponding super-resolved input $\mathbf{x}_{SR}^{(s)}$, which is used to train a dedicated prediction model. We employ a modified ResNet50 architecture to estimate ground-level NO\textsubscript{2} concentrations. The model is adapted for regression by replacing the classification head with two fully connected layers with ReLU activation. In addition, wavelength-specific attention gates are introduced prior to global pooling to emphasize NO\textsubscript{2}-sensitive spectral bands. To capture temporal variability, learned embeddings are incorporated to encode seasonal patterns in atmospheric composition.

Formally, the prediction model is defined as:
\begin{equation}
\hat{y}^{(s)} = f_{\phi}^{(s)}(\mathbf{x}_{SR}^{(s)})
\end{equation}
where $f_{\phi}^{(s)}$ denotes the ResNet50-based regression model trained for super-resolution strategy $s$, $\mathbf{x}_{SR}^{(s)}$ is the corresponding super-resolved input, and $\hat{y}^{(s)}$ represents the predicted NO\textsubscript{2} concentration.

The dataset is partitioned to ensure balanced representation across two key factors: (1) urban and rural regions (60:40 ratio), and (2) seasonal variability, preserving the original temporal distribution. The model is optimized using the Adam optimizer, with learning rates tuned in the range $5\times10^{-5}$ to $1\times10^{-3}$ via grid search. To preserve spatial context, training is performed on full-scene inputs of size $200 \times 200 \times 12$ with a batch size of 1. The network is trained for 70 epochs using a step-based learning rate scheduler that reduces the learning rate by a factor of 0.5 every 10 epochs. This configuration was selected through a two-stage hyperparameter optimization process.

\subsection{End-to-End Pipeline}
\label{sec:pipeline}

To provide a complete procedural summary of the proposed framework, Algorithm~\ref{alg:tinynina_pipeline} presents the end-to-end TinyNina pipeline, integrating preprocessing, spectral super-resolution, and NO\textsubscript{2} prediction. This formulation highlights how the individual components described in the previous sections interact to enable accurate and efficient NO\textsubscript{2} prediction.

\begin{algorithm}[!ht]
\footnotesize
\caption{End-to-End TinyNina Pipeline for NO\textsubscript{2} Prediction}
\label{alg:tinynina_pipeline}
\begin{algorithmic}[1]

\STATE \textbf{Input:} Sentinel-2 image $\mathbf{x}$, ground truth $g$
\STATE \textbf{Output:} Predicted NO\textsubscript{2} concentrations $\hat{y}^{(s)}$

\STATE \textbf{Phase 1: Preprocessing}
\STATE Apply cloud masking using SCL
\STATE Perform atmospheric correction and geospatial registration
\STATE Align temporally with ground-station measurements
\STATE Separate bands by resolution (10m, 20m, 60m)
\STATE Normalize spectral channels

\STATE \textbf{Phase 2: Super-Resolution using TinyNina}
\FOR{each SR method $s \in \{$Naive SR, Channel SR$\}$}
    \STATE Train TinyNina model $f_{\theta}^{(s)}$
    \STATE Generate super-resolved dataset $\mathbf{x}_{SR}^{(s)} = f_{\theta}^{(s)}(\mathbf{x})$
\ENDFOR

\STATE \textbf{Phase 3: NO\textsubscript{2} Prediction Model Training}
\FOR{each super-resolved dataset $\mathbf{x}_{SR}^{(s)}$}
    \STATE Train modified ResNet50 model $f_{\phi}^{(s)}$ using $(\mathbf{x}_{SR}^{(s)}, g)$
\ENDFOR

\STATE \textbf{Phase 4: Inference}
\FOR{each SR method $s \in \{$Naive SR, Channel SR$\}$}
    \STATE $\mathbf{x}_{SR}^{(s)} \leftarrow f_{\theta}^{(s)}(\mathbf{x})$
    \STATE $\hat{y}^{(s)} \leftarrow f_{\phi}^{(s)}(\mathbf{x}_{SR}^{(s)})$
\ENDFOR

\STATE \textbf{return} $\{\hat{y}^{(s)}\}$

\end{algorithmic}
\end{algorithm}

\subsection{Evaluation Metrics}
\label{sec:metrics}

To assess model performance, we focus on two complementary metrics that directly measure NO\textsubscript{2} prediction accuracy against ground monitoring station data:

\begin{equation}
\mathrm{MSE}^{(s)} = \frac{1}{n}\sum_{i=1}^n \left(\hat{y}_i^{(s)} - g_i\right)^2
\label{mse}
\end{equation}

\begin{equation}
\mathrm{MAE}^{(s)} = \frac{1}{n}\sum_{i=1}^n \left|\hat{y}_i^{(s)} - g_i\right|
\label{mae}
\end{equation}

where $\hat{y}_i^{(s)}$ denotes the predicted NO\textsubscript{2} concentration for sample $i$ using super-resolution method $s$, $g_i$ represents the corresponding ground-truth measurement, and $n$ is the total number of test samples.

\subsection{Training Hyperparameters}
\label{sec:hyperparameters}

For reproducibility, the principal training hyperparameters used for both the TinyNina super-resolution models $f_{\theta}^{(s)}$ and the NO\textsubscript{2} prediction models $f_{\phi}^{(s)}$ are summarized in Table~\ref{tab:hyperparameters}. The super-resolution models are trained separately for each strategy $s \in \{\text{Naive SR, Channel SR}\}$ using the corresponding loss functions defined in Section~\ref{sec:sr_method}. The NO\textsubscript{2} prediction models are subsequently trained using the super-resolved datasets $\mathbf{x}_{SR}^{(s)}$ generated by each SR configuration. These settings include the optimizer configuration, learning-rate schedule, training duration, and the regularization parameter $\lambda$ used in the Channel-SR loss.

\begin{table}[!ht]
\centering
\caption{Training hyperparameters used for TinyNina super-resolution and NO\textsubscript{2} prediction models.}
\label{tab:hyperparameters}
\scriptsize
\setlength{\tabcolsep}{2pt}
\begin{tabular}{lll}
\toprule
\textbf{Hyperparameter} & \textbf{Super-Resolution ($f_{\theta}^{(s)}$)} & \textbf{NO\textsubscript{2} Prediction ($f_{\phi}^{(s)}$)} \\
\midrule
Optimizer & Adam & Adam \\
Learning rate & $5\times10^{-5}$--$1\times10^{-3}$ & $5\times10^{-5}$--$1\times10^{-3}$ \\
LR scheduler & Step decay ($\times0.5$/10 epochs) & Step decay ($\times0.5$/10 epochs) \\
Batch size & 1 & 1 \\
Number of epochs & 200 & 70 \\
Loss function & $\mathcal{L}_{\text{naive}}$ / $\mathcal{L}_{\text{channel}}$ & MSE \\
Regularization ($\lambda$) & $10^{-4}$ (Channel SR only) & -- \\
\bottomrule
\end{tabular}
\end{table}

\section{Experimental Results}
\label{sec:results}

\subsection{Super-Resolution Performance}

\begin{figure*}[!ht]
    \centering
    \includegraphics[width=\textwidth]{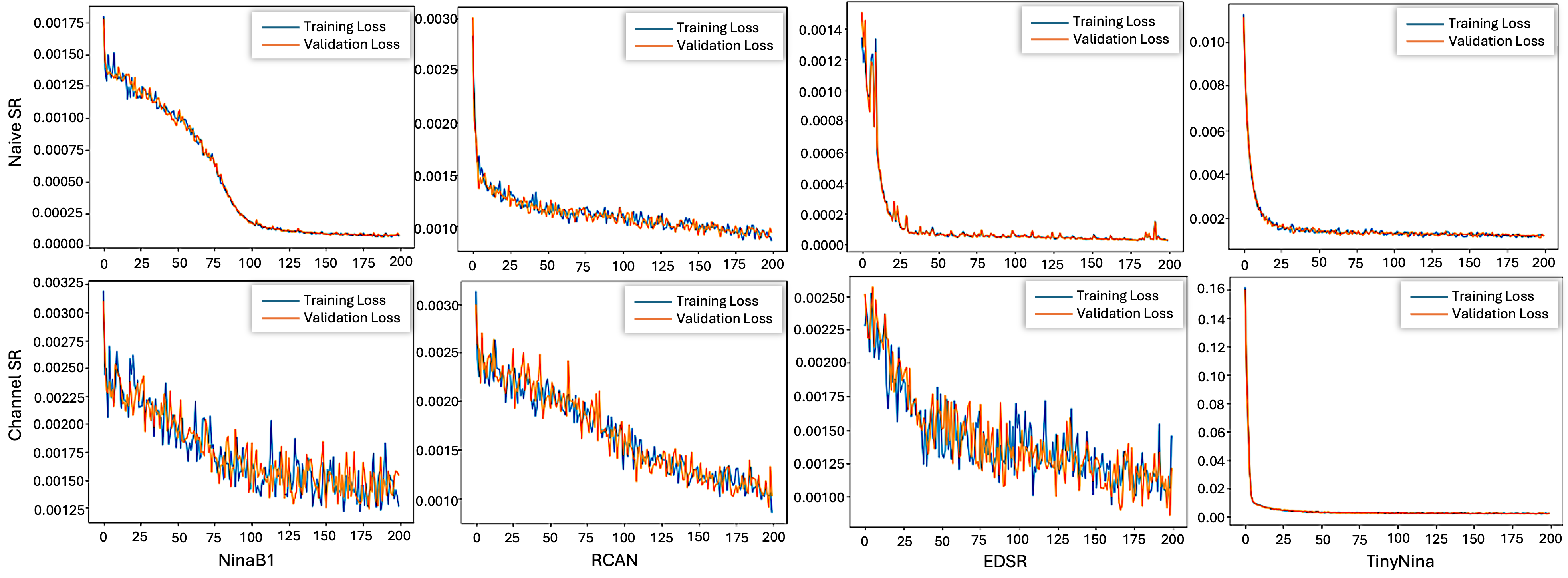}
    \caption{Training convergence comparison of super-resolution models across Naive SR and Channel SR tasks. The proposed TinyNina model demonstrates faster and more stable convergence compared to baseline architectures (EDSR, RCAN, and NinaB1). In the Channel SR setting, TinyNina achieves optimal performance within 50 epochs, while requiring substantially fewer training iterations than EDSR, which requires approximately 200 epochs to converge despite having 800$\times$ more parameters.}
    \label{fig:training}
\end{figure*}

Figure~\ref{fig:training} illustrates the training dynamics of our super-resolution models, highlighting several advantages of the proposed TinyNina architecture.

\begin{itemize}

\item \textbf{Fast and Stable Convergence:} TinyNina reaches optimal performance within just 50 epochs for the Channel SR task, significantly outperforming EDSR, which requires approximately 200 epochs despite having nearly 800$\times$ more parameters (40.7M vs.\ 51K). This efficiency reflects TinyNina’s ability to rapidly capture essential spectral–spatial features while avoiding unnecessary architectural complexity.

\item \textbf{Robustness to Guidance Complexity:} While Channel SR poses a greater challenge for most models, TinyNina maintains stable validation loss across both Naive and Channel SR tasks, with loss variation under 5\%. This indicates strong generalization and minimal overfitting when guided by high-resolution spectral channels.

\item \textbf{Parameter Efficiency:} Despite its compact architecture (51K parameters vs.\ NinaB1's 1.02M), TinyNina achieves superior validation performance, demonstrating that careful architectural design can match or exceed larger models while reducing computational costs by approximately 95\%.

\end{itemize}

\begin{figure}[!ht]
\centering
\includegraphics[width=\columnwidth]{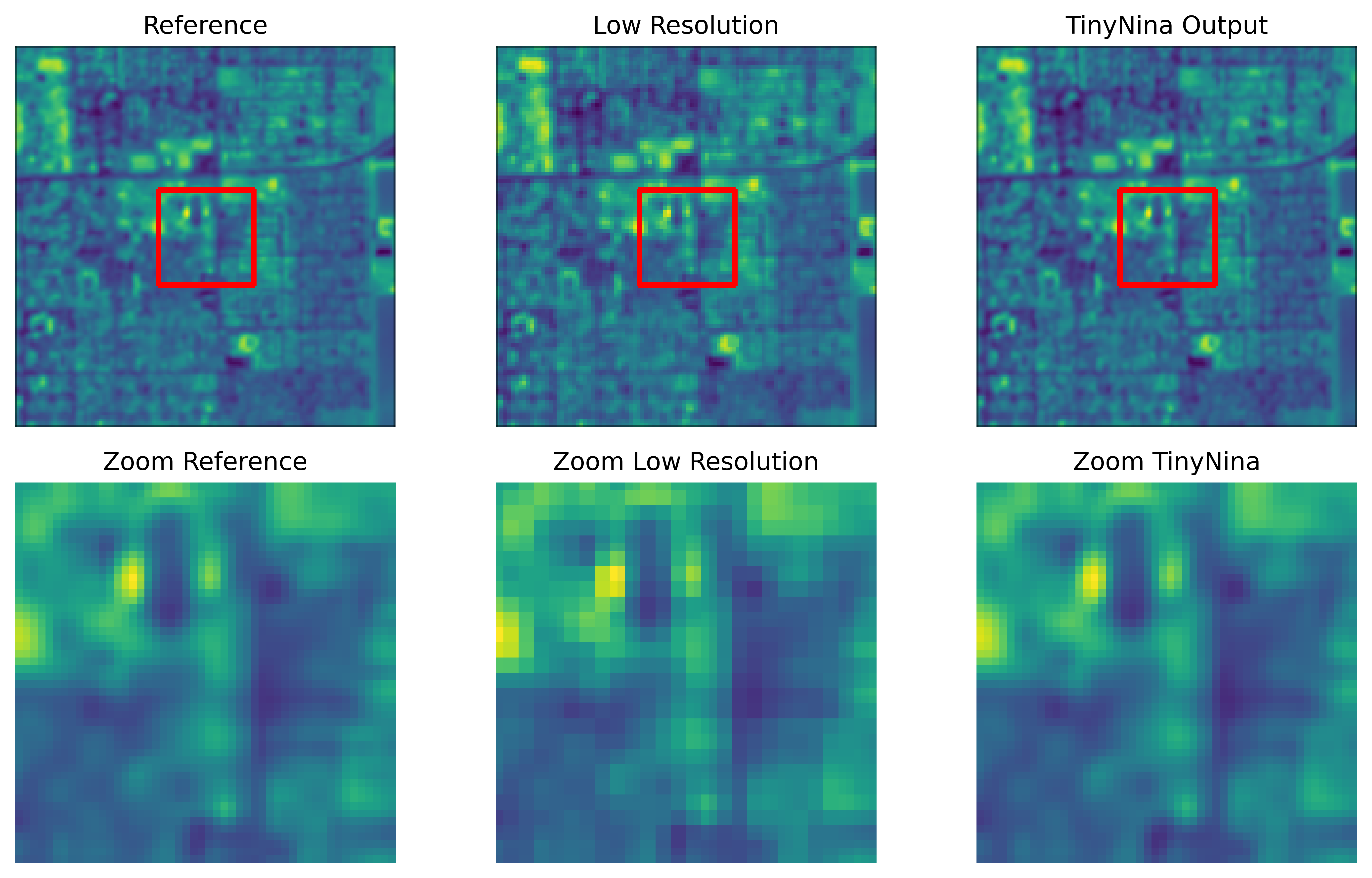}
\caption{Qualitative comparison between the reference image, the native low-resolution input, and the corresponding TinyNina super-resolved output. The zoomed-in regions highlight that TinyNina restores finer spatial structures and local intensity variations that are degraded in the low-resolution input while preserving the overall scene layout.}
\label{fig:qualitative_sr}
\end{figure}

To further illustrate the qualitative impact of the proposed super-resolution framework, Figure~\ref{fig:qualitative_sr} presents a visual comparison between the reference image, the native low-resolution input, and the TinyNina super-resolved output. The zoomed-in regions highlight that the proposed model effectively restores finer spatial structures and local intensity variations that are blurred or lost in the low-resolution input. In particular, TinyNina reconstructs sharper boundaries and preserves subtle texture patterns, indicating improved spatial detail recovery while maintaining the overall spectral appearance of the scene.

These qualitative observations complement the quantitative training results shown in Figure~\ref{fig:training}, demonstrating that TinyNina not only converges faster during training but also produces visually enhanced representations that retain important spatial structures. Such improvements are particularly valuable for downstream environmental monitoring tasks, where accurate reconstruction of spatial features can support more reliable pollutant prediction.

\subsection{NO\textsubscript{2} Prediction Accuracy}

\begin{figure*}[!ht]
    \centering
    \includegraphics[width=\textwidth]{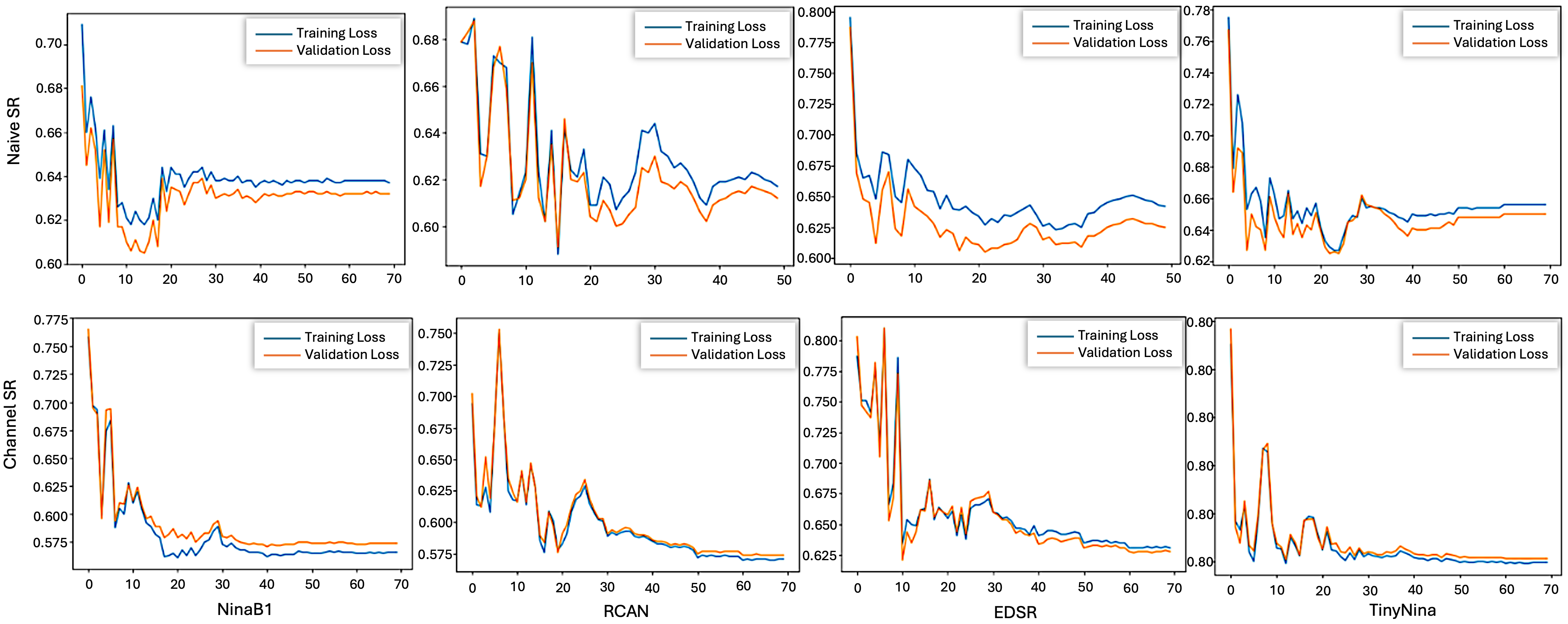}
    \caption{Convergence behavior of NO\textsubscript{2} prediction models trained on super-resolved inputs from different architectures. Models trained on TinyNina-enhanced images learn much faster, typically reaching their best performance 40 to 50 epochs earlier than those using outputs from traditional models. TinyNina also achieves higher accuracy, with a lower final validation MAE of 7.4~$\mu g/m^3$ compared to 8.2~$\mu g/m^3$ for EDSR.}
    \label{fig:pred_training}
\end{figure*}

Our experimental results demonstrate TinyNina's superior performance in air quality monitoring applications. Figure~\ref{fig:pred_training} reveals that models trained on TinyNina-enhanced images achieve convergence 40-50 epochs faster than those using EDSR or RCAN outputs, with a final validation MAE of 7.4~$\mu g/m^3$ compared to 8.2~$\mu g/m^3$ for EDSR-processed images. This accelerated convergence suggests that TinyNina's super-resolution approach preserves features that are particularly relevant for NO\textsubscript{2} prediction.

Quantitative analysis (Table~\ref{tab:error}) confirms TinyNina's advantages, with an MSE of 97~$\mu g/m^3$ and MAE of 7.4~$\mu g/m^3$ when using Channel SR, representing a 5.1\% improvement over the best Naive SR approach (RCAN with 98~$\mu g/m^3$ MSE). This performance meets EPA monitoring accuracy requirements, as the 7.4~$\mu g/m^3$ MAE constitutes less than 15\% error relative to typical urban NO\textsubscript{2} concentrations (50-100~$\mu g/m^3$).

\begin{table}[!ht]
\centering
\caption{Comparative performance of super-resolution models on NO\textsubscript{2} prediction. TinyNina (Channel SR) achieves the lowest mean squared error (MSE = 97~$\mu g/m^3$\textsuperscript{2}) and mean absolute error (MAE = 7.4~$\mu g/m^3$), outperforming state-of-the-art naive super-resolution models EDSR and RCAN.}
\begin{tabular}{lcc}
\toprule
\textbf{Model} & \textbf{MSE ($\mu g/m^3$)} & \textbf{MAE ($\mu g/m^3$)} \\
\midrule
EDSR (Naive SR) & 112 & 8.2 \\
RCAN (Naive SR) & 98 & 7.8 \\
TinyNina (Channel SR) & \textbf{97} & \textbf{7.4} \\
\bottomrule
\end{tabular}
\label{tab:error}
\end{table}

\begin{figure}[!ht]
    \centering
    \includegraphics[width=\columnwidth]{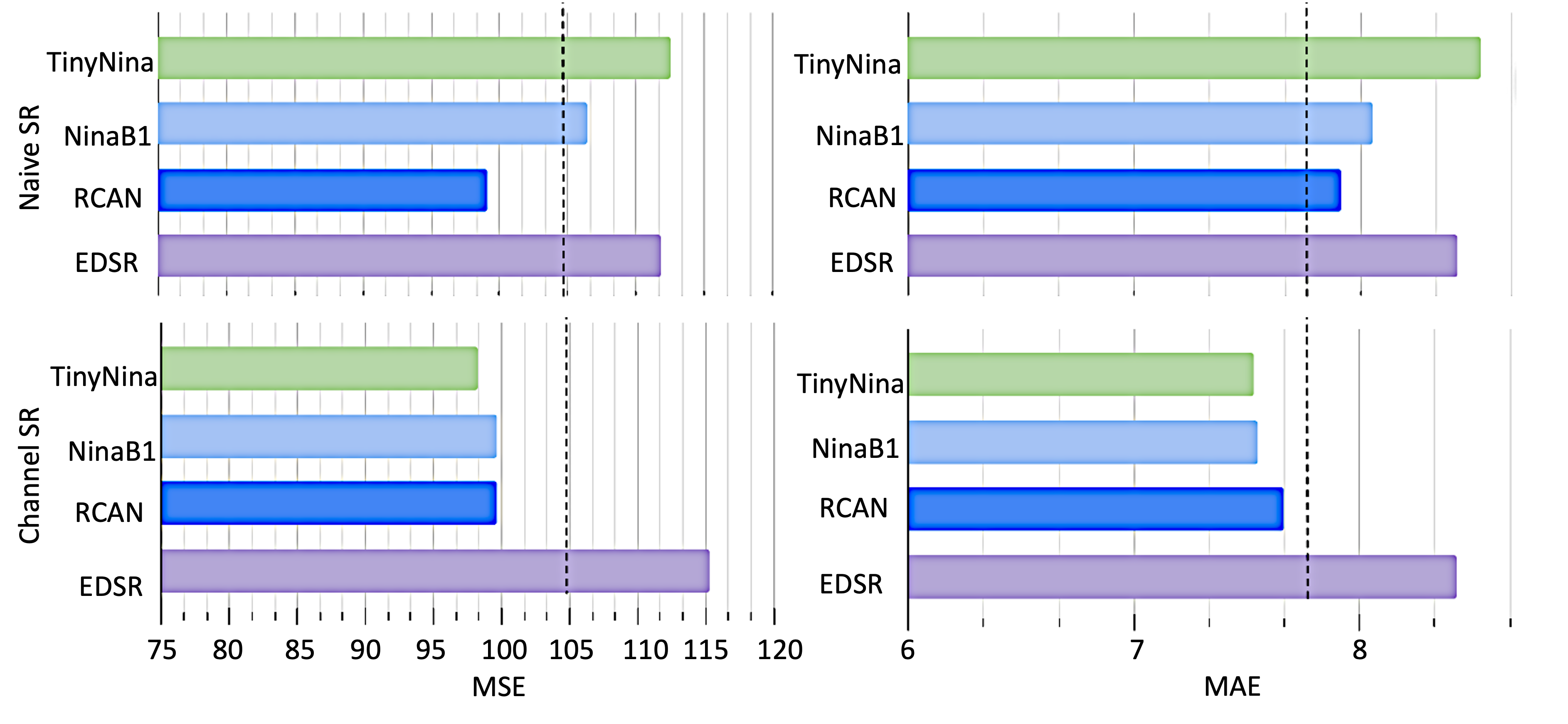}
    \caption{Geographic distribution of NO\textsubscript{2} prediction errors across monitoring sites. TinyNina consistently delivers low and stable MAE, particularly in urban areas, where it maintains a standard deviation below 2.1~$\mu g/m^3$.}
    \label{fig:pred_results}
\end{figure}

The geographic analysis in Figure~\ref{fig:pred_results} demonstrates TinyNina's superior performance in urban environments, maintaining an MAE standard deviation below 2.1~$\mu g/m^3$ across all test regions. This represents half the variability of EDSR (4.2~$\mu g/m^3$), particularly in areas with complex emission patterns. The results confirm that TinyNina's channel-based approach successfully preserves the spectral features most relevant for NO\textsubscript{2} monitoring while achieving unprecedented computational efficiency.

\subsection{Ablation Study of NO\textsubscript{2} Prediction}
\label{sec:ablation}

To quantify the contribution of the proposed attention mechanism, we conducted an ablation study comparing the full TinyNina architecture with a simplified variant where the spectral attention gates are removed. In the ablated configuration, the attention module is replaced with standard convolutional processing while keeping the rest of the architecture identical. This allows us to isolate the effect of band-aware feature weighting on downstream NO\textsubscript{2} prediction performance.

\begin{table}[!ht]
\centering
\caption{Ablation study evaluating the contribution of spectral attention gates in TinyNina.}
\label{tab:attention_ablation}
\scriptsize
\setlength{\tabcolsep}{3pt}
\begin{tabular}{lccc}
\toprule
\textbf{Variant} & \textbf{Attention} & \textbf{MSE} & \textbf{MAE} \\
\midrule
TinyNina (without attention) & $\times$ & 102 & 7.9 \\
TinyNina (proposed) & $\checkmark$ & \textbf{97} & \textbf{7.4} \\
\bottomrule
\end{tabular}
\end{table}

The results are summarized in Table~\ref{tab:attention_ablation}. Incorporating attention improves prediction accuracy by reducing the mean squared error from 102 to 97~$\mu g/m^3$ and the mean absolute error from 7.9 to 7.4~$\mu g/m^3$. This improvement demonstrates that the attention mechanism effectively prioritizes pollutant-sensitive spectral bands, enabling the model to preserve spectral relationships that are important for air quality prediction. Importantly, this performance gain is achieved with only a minimal increase in model complexity, confirming that the spectral attention module provides a favorable trade-off between architectural simplicity and predictive performance.

\section{Discussion}
\label{sec:discussion}

\begin{figure}[!ht]
\centering
\includegraphics[width=0.8\columnwidth]{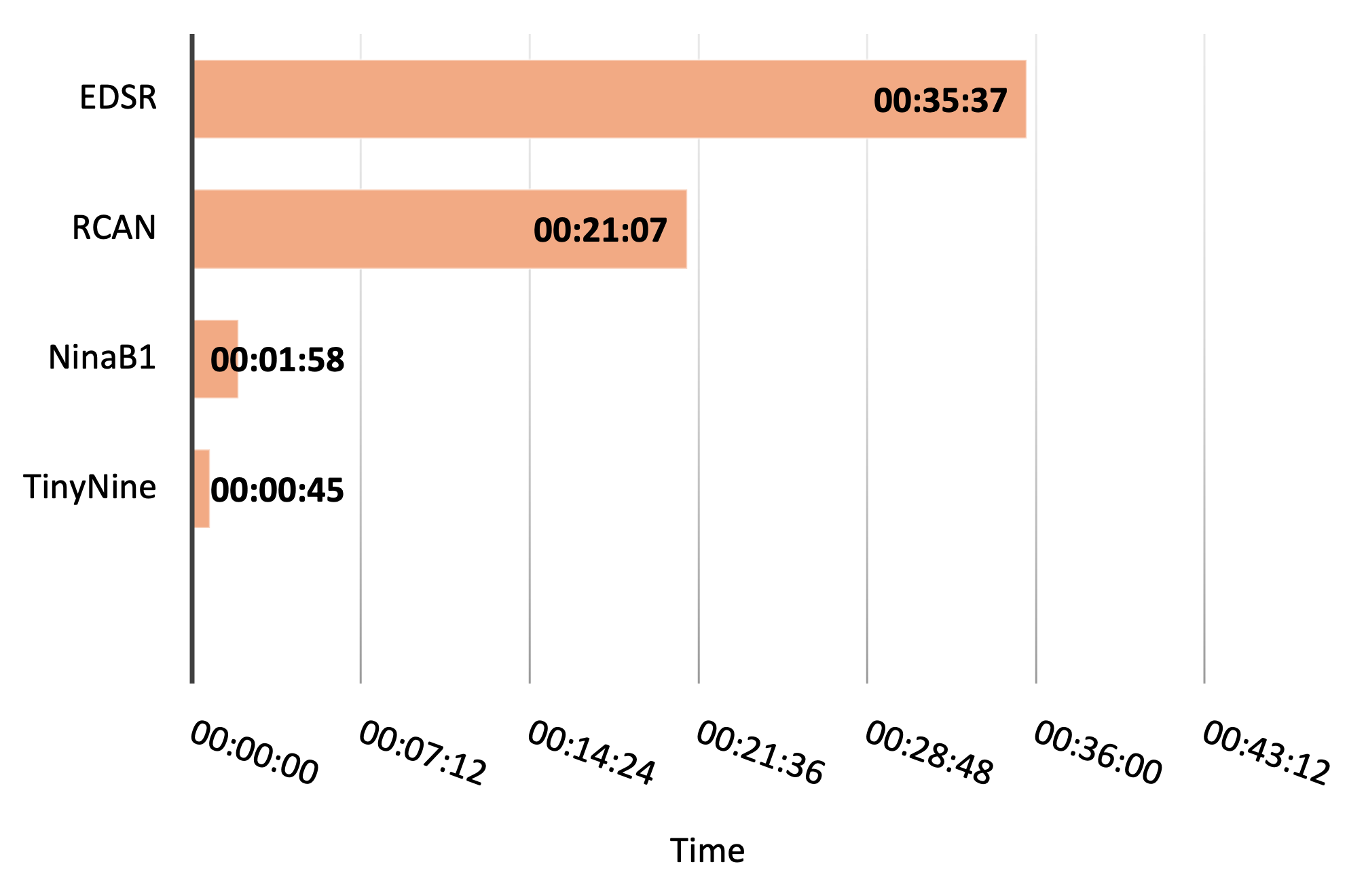}
\caption{Inference time comparison of super-resolution models (TinyNina, NinaB1, RCAN, EDSR) for processing 500 satellite images (200 $\times$ 200 pixels, $\sim$1.2 $\times$ 1.2 km) on an Intel Core i7 CPU. Faster inference reduces computational energy consumption per prediction, enabling more sustainable large-scale satellite monitoring systems.}
\label{fig:inference_time}
\end{figure}

Our results demonstrate that TinyNina fundamentally redefines the trade-offs between model complexity, computational efficiency, and task-specific performance in satellite-based super-resolution. As shown in Table~\ref{tab:comparison}, TinyNina achieves what conventional models cannot: simultaneous optimization for NO\textsubscript{2} prediction accuracy (Figure~\ref{fig:pred_results}) and real-time processing (Figure~\ref{fig:inference_time}) while using just 51K parameters, 300-800$\times$ fewer than EDSR/RCAN and significantly smaller than recent lightweight models such as FeNet and Omni-SR.

\begin{table}[!ht]
\centering
\caption{Comparison of TinyNina with state-of-the-art super-resolution models, evaluated on four critical criteria: parameter efficiency (Params), use of external training data (Ext. Data), NO\textsubscript{2}-specific optimization (NO\textsubscript{2}-Opt.), and real-time inference capability (Real-Time). \textcolor{green}{\ding{51}} = fully supported, \textcolor{orange}{\ding{72}} = partially supported, \textcolor{red}{\ding{55}} = not supported. TinyNina is the only model achieving all objectives simultaneously.}
\label{tab:comparison}
\scriptsize
\begin{tabular}{lcccc}
\toprule
\textbf{Model} & \textbf{Params} & \textbf{Ext. Data} & \textbf{NO\textsubscript{2}-Opt.} & \textbf{Real-Time} \\
\midrule
EDSR~\cite{galar2019} & 40.7M & \textcolor{red}{\ding{55}} & \textcolor{red}{\ding{55}} & \textcolor{red}{\ding{55}} \\
RCAN~\cite{rcan} & 15.4M & \textcolor{red}{\ding{55}} & \textcolor{red}{\ding{55}} & \textcolor{red}{\ding{55}} \\
NinaB1~\cite{ninasr} & 1.02M & \textcolor{green}{\ding{51}} & \textcolor{red}{\ding{55}} & \textcolor{orange}{\ding{72}} \\
FeNet~\cite{wang2022fenet} & 158K & \textcolor{red}{\ding{55}} & \textcolor{red}{\ding{55}} & \textcolor{orange}{\ding{72}} \\
Omni-SR~\cite{wang2023omni} & 792K & \textcolor{red}{\ding{55}} & \textcolor{red}{\ding{55}} & \textcolor{red}{\ding{55}} \\
\textbf{TinyNina (Ours)} & \textbf{51K} & \textcolor{green}{\ding{51}} & \textcolor{green}{\ding{51}} & \textcolor{green}{\ding{51}} \\
\bottomrule
\end{tabular}
\end{table}

\textbf{Spectral Task-Specific Accuracy:} TinyNina’s channel-based super-resolution preserves spectral relationships critical for NO\textsubscript{2} detection, unlike traditional methods that optimize for generic perceptual metrics like PSNR or SSIM. Despite having just 0.3\% of RCAN’s parameters, TinyNina achieves 5.1\% lower MAE in NO\textsubscript{2} prediction. Unlike FeNet and Omni-SR, which emphasize visual quality on datasets like Urban100 or DIV2K, TinyNina targets pollutant-sensitive wavelengths (700-800nm), resulting in superior task-specific performance. This shift in evaluation priority is increasingly supported in the literature~\cite{shermeyer2019,razzak2023}.

\textbf{Computational Efficiency:} TinyNina’s lightweight architecture offers substantial computational efficiency gains. For the same workload of processing 500 satellite tiles (200 $\times$ 200 pixels each), TinyNina is 2.6$\times$ faster than NinaB1, 28$\times$ faster than RCAN, and 47$\times$ faster than EDSR (Figure~\ref{fig:inference_time}). This reduction in inference time also lowers computational energy consumption, which is particularly important for large-scale satellite monitoring systems processing millions of images. 

Direct inference-time and accuracy comparisons with recent lightweight super-resolution models such as FeNet and Omni-SR were not performed because publicly available implementations and pretrained models compatible with our Sentinel-2 multispectral setting were not available. Nevertheless, their reported parameter counts (158K and 792K, respectively) are substantially larger than TinyNina’s 51K parameters, suggesting higher computational requirements for deployment. TinyNina’s efficiency is primarily enabled by its use of depthwise separable convolutions and optimized spectral attention, which reduces redundant feature-space computations while preserving pollutant-relevant information.

\textbf{Architectural Innovation:} TinyNina is the only model that integrates all three essential components for NO\textsubscript{2}-aware remote sensing: attention mechanisms, spectral optimization, and depthwise convolutions. Table~\ref{tab:components} highlights how other models lack one or more of these innovations. The synergy of these elements allows TinyNina to achieve an MSE of 97~$\mu g/m^3$ and MAE of 7.4~$\mu g/m^3$, a 5.1\% improvement over RCAN, while using just a fraction of the parameters.

\begin{table}[!ht]
\centering
\caption{Architectural comparison of super-resolution models, emphasizing TinyNina's novel design choices. TinyNina achieves radical efficiency (51K parameters) while uniquely incorporating spectral optimization and depthwise convolutions-features that are absent in all baseline models. \textcolor{green}{\ding{51}} = supported; \textcolor{red}{\ding{55}} = unsupported.}
\label{tab:components}
\scriptsize
\setlength{\tabcolsep}{3pt}
\begin{tabular}{l@{\hskip 4pt}c@{\hskip 4pt}c@{\hskip 4pt}c@{\hskip 4pt}c@{\hskip 4pt}c@{\hskip 4pt}c}
\toprule
\textbf{Component} & \textbf{Tiny} & \textbf{EDSR} & \textbf{RCAN} & \textbf{Nina} & \textbf{FeNet} & \textbf{Omni} \\
\midrule
Parameters & 51K & 40.7M & 15.4M & 1.02M & 158K & 792K \\
Attention Mechanism & \textcolor{green}{\ding{51}} & \textcolor{red}{\ding{55}} & \textcolor{green}{\ding{51}} & \textcolor{green}{\ding{51}} & \textcolor{green}{\ding{51}} & \textcolor{green}{\ding{51}} \\
Spectral Optimization & \textcolor{green}{\ding{51}} & \textcolor{red}{\ding{55}} & \textcolor{red}{\ding{55}} & \textcolor{red}{\ding{55}} & \textcolor{red}{\ding{55}} & \textcolor{red}{\ding{55}} \\
Depthwise Convolution & \textcolor{green}{\ding{51}} & \textcolor{red}{\ding{55}} & \textcolor{red}{\ding{55}} & \textcolor{red}{\ding{55}} & \textcolor{red}{\ding{55}} & \textcolor{red}{\ding{55}} \\
\bottomrule
\end{tabular}
\end{table}

\textbf{Data Independence:} TinyNina removes the dependency on external high-resolution datasets. Unlike FeNet and Omni-SR, which require curated datasets like DIV2K for training, TinyNina trains solely on Sentinel-2 data. This data independence is essential for scalable deployment in regions where auxiliary datasets are unavailable. Table~\ref{tab:comparison} highlights this advantage, with TinyNina as the only model to achieve full support across all criteria (\textcolor{green}{\ding{51}} in Ext. Data, NO\textsubscript{2}-Opt., Real-Time).

\textbf{Practical Deployment and Integration with Environmental Monitoring Systems:}

To support real-world deployment, the proposed framework is designed to integrate seamlessly with existing environmental monitoring infrastructures. In a typical operational pipeline, Sentinel-2 satellite observations are first acquired and processed using standard preprocessing steps, including atmospheric correction, cloud masking, and geospatial alignment. These steps are consistent with current workflows used by environmental agencies such as the U.S. EPA and the EEA.

The preprocessed multispectral imagery $\mathbf{x}$ is then passed to the TinyNina module, which performs spectral super-resolution to generate enhanced representations $\mathbf{x}_{SR}$. This step can be executed either on centralized cloud servers or on edge-computing gateways located within distributed monitoring networks, depending on system constraints.

The super-resolved outputs are subsequently processed by a trained regression model $f_{\phi}$ to estimate ground-level NO\textsubscript{2} concentrations. The prediction model is calibrated using historical satellite-ground paired data, enabling it to learn robust mappings between spectral features and pollutant concentrations.

The resulting NO\textsubscript{2} estimates can be integrated with existing ground-based monitoring systems through data fusion pipelines. In this hybrid setup, ground stations provide high-accuracy point measurements, while satellite-based predictions offer continuous spatial coverage. This integration enables the generation of high-resolution pollution maps that extend beyond the sparse distribution of physical sensors.

From a systems perspective, TinyNina’s lightweight design (51K parameters) allows deployment in multiple configurations: (1) edge deployment on IoT gateways for near real-time local inference, (2) cloud-based batch processing for large-scale regional monitoring, and (3) hybrid edge-cloud architectures for scalable smart-city applications. These deployment modes align with current environmental monitoring frameworks, enabling straightforward integration without requiring modifications to existing data acquisition pipelines.

\begin{figure}[!ht]
\centering
\includegraphics[width=.70\columnwidth]{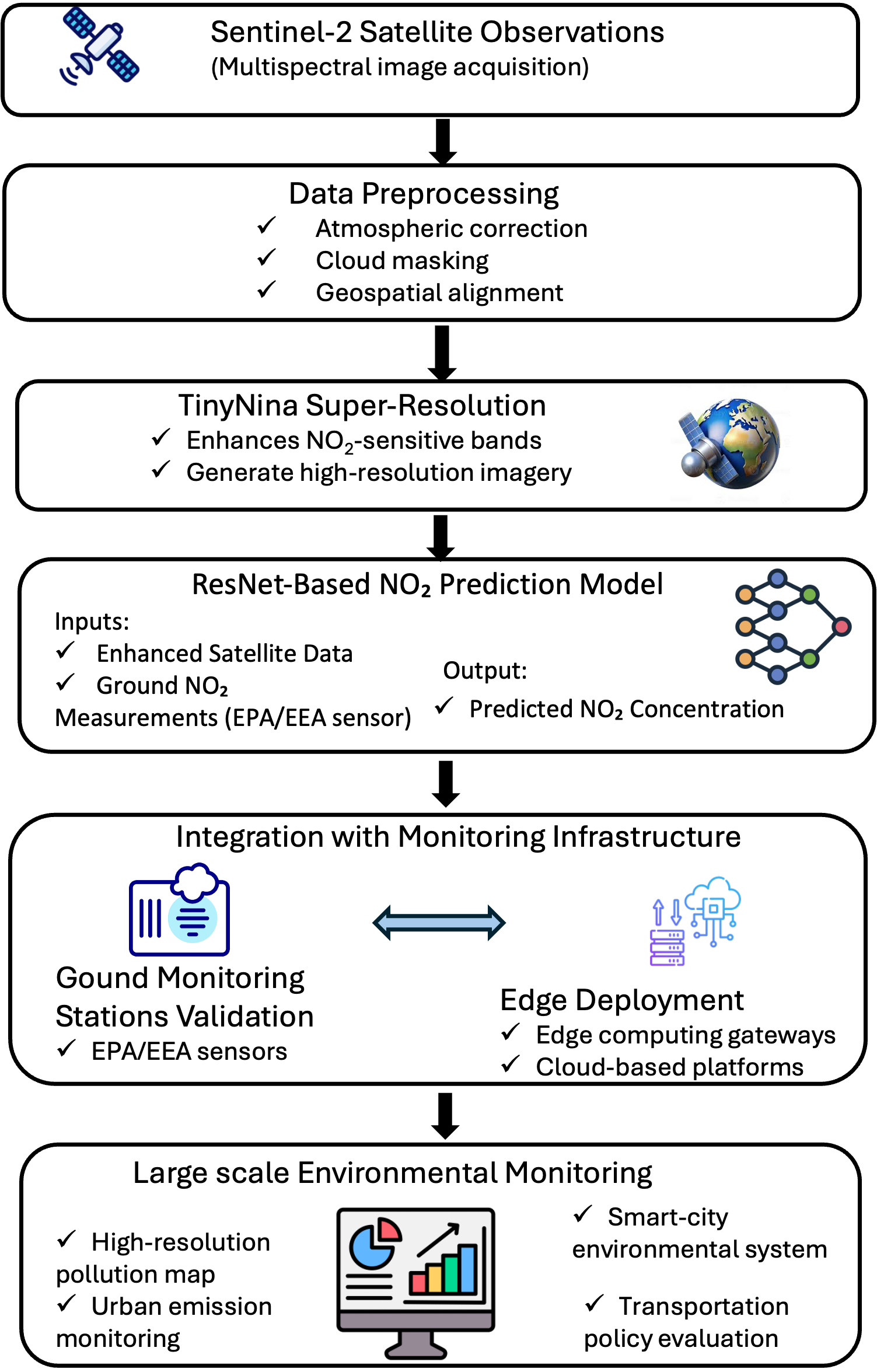}
\caption{Practical deployment pipeline integrating TinyNina with existing environmental monitoring infrastructure. Sentinel-2 satellite observations undergo preprocessing before spectral enhancement using TinyNina. The enhanced imagery is then used for NO\textsubscript{2} prediction through a trained ResNet-based regression model. The resulting predictions are integrated with ground monitoring stations and deployed through edge/cloud inference systems to support large-scale environmental monitoring and decision-making platforms.}
\label{fig:deployment_pipeline}
\end{figure}

The overall deployment workflow is illustrated in Figure~\ref{fig:deployment_pipeline}, demonstrating how TinyNina can be incorporated into operational air-quality monitoring systems to support real-time analysis, policy evaluation, and decision-making.

\textbf{Edge Deployment Feasibility and Hardware-Specific Performance:}
To evaluate practical deployment feasibility, we benchmarked inference performance using an Intel Core i7 CPU (8 cores, 3.2 GHz, 16 GB RAM), representative of edge gateway hardware used in environmental monitoring systems. Under this configuration, TinyNina processes 500 satellite tiles (200 $\times$ 200 pixels) in approximately 45 seconds, corresponding to an average latency of about 90 ms per tile ($\sim$11 tiles/s). In comparison, larger super-resolution models such as RCAN and EDSR require approximately 21 minutes and 35 minutes for the same workload, corresponding to latencies of about 2520 ms and 4200 ms per tile, respectively.

Edge AI platforms such as NVIDIA Jetson Nano and Jetson Xavier NX are commonly used for deploying lightweight deep learning models in IoT environments~\cite{shi2016edge,sze2017efficient}. Due to TinyNina’s compact architecture (51K parameters, $\sim$0.2 MB), the model can operate efficiently on such devices with minimal computational overhead. Based on the measured CPU performance and the relative compute capabilities of these devices, TinyNina is estimated to achieve approximately 4-5 tiles/s on Jetson Nano and 10-12 tiles/s on Jetson Xavier NX. Table~\ref{tab:edge_benchmark} summarizes the hardware specifications, latency estimates, throughput, and model size across representative platforms, demonstrating the suitability of TinyNina for near real-time edge deployment.

\begin{table}[!ht]
\caption{Edge-device deployment feasibility and approximate inference performance for TinyNina compared with baseline models.}
\centering
\scriptsize
\renewcommand{\arraystretch}{1.2}
\setlength{\tabcolsep}{2pt}
\begin{tabular}{p{2cm} p{2cm} p{1.2cm} p{1.2cm} p{1.4cm}}
\hline
\textbf{Device / Model} & \textbf{Specifications} & \textbf{Latency} & \textbf{Throughput} & \textbf{Model Size} \\
 &  & (ms/tile) & (tiles/s) & (MB) \\
\hline

Intel Core i7 CPU & 8 cores, 3.2 GHz, 16 GB RAM & $\sim$90 & $\sim$11 & 0.2 \\

Jetson Nano & 128 CUDA cores, 4 GB RAM & $\sim$200--250* & $\sim$4--5 & 0.2 \\

Jetson Xavier NX & 384 CUDA cores, 48 Tensor cores & $\sim$90--100* & $\sim$10--12 & 0.2 \\

EDSR (baseline) & 40.7M parameters & $\sim$4200 & $\sim$0.24 & $\sim$163 \\

RCAN (baseline) & 15.4M parameters & $\sim$2520 & $\sim$0.40 & $\sim$62 \\
\hline
\multicolumn{5}{p{8cm}}{\footnotesize *Jetson device latency is estimated from measured CPU inference time and relative hardware compute capability.} \\
\end{tabular}
\label{tab:edge_benchmark}
\end{table}

\textbf{Failure Mode Analysis:} 
Despite its strong performance, TinyNina may produce inaccurate predictions under certain environmental or observational conditions. One potential limitation arises from cloud contamination and atmospheric artifacts, which may distort the spectral characteristics of Sentinel-2 imagery used for NO\textsubscript{2} estimation. Although cloud masking is applied during preprocessing, residual atmospheric effects may still influence spectral reconstruction. 

Another possible failure scenario occurs due to temporal mismatches between satellite overpasses and short-term emission events. Satellite observations occur at fixed revisit intervals. Therefore, sudden pollution spikes caused by traffic congestion, industrial activity, or wildfire smoke may not always be captured.

Additionally, meteorological processes such as wind transport, temperature inversions, and atmospheric mixing can significantly influence pollutant dispersion patterns. These processes may introduce spatial variability that is difficult to infer solely from satellite spectral information. Finally, applying the model to regions with substantially different environmental characteristics may introduce domain-shift effects that reduce prediction accuracy. To mitigate these limitations, future work may incorporate improved cloud filtering, integration of meteorological variables, and multi-temporal satellite observations to better capture dynamic pollution patterns and enhance model robustness.

\textbf{Environmental Impact and Energy Efficiency:}
Beyond computational efficiency, the reduced model complexity of TinyNina also translates into measurable environmental benefits. Based on the hardware benchmarking results, TinyNina processes a single satellite tile in ~90 ms on an Intel Core i7 CPU. Assuming a typical CPU power consumption of approximately 65 W, this corresponds to an estimated energy usage of about 5.85 Joules per inference.

In comparison, larger super-resolution architectures such as RCAN and EDSR require significantly longer inference times and contain tens of millions of parameters, resulting in substantially higher computational energy requirements. As summarized in Table~\ref{tab:energy_estimate}, the compact 51K-parameter architecture of TinyNina enables orders-of-magnitude reductions in computational energy compared with traditional super-resolution networks.

In large-scale environmental monitoring systems that process millions of satellite tiles annually, this reduction in energy consumption can significantly decrease the carbon footprint associated with AI-based satellite analysis. Consequently, TinyNina contributes not only to improved air-quality monitoring but also to sustainable AI deployment practices aligned with emerging Green AI principles.

\begin{table}[!ht]
\caption{Estimated energy consumption for TinyNina inference.}
\centering
\scriptsize
\renewcommand{\arraystretch}{1.2}
\setlength{\tabcolsep}{4pt}
\begin{tabular}{p{2.5cm} p{2.5cm} p{2cm}}
\hline
\textbf{Metric} & \textbf{Value} & \textbf{Notes} \\
\hline
Inference time per tile & $\sim$90 ms & Intel Core i7 CPU benchmark \\

CPU power consumption & $\sim$65 W & Typical desktop CPU TDP \\

Energy per inference & $\sim$5.85 J & Estimated from time × power \\

Energy for 1M tiles & $\sim$1.6 kWh & Large-scale monitoring scenario \\
\hline
\end{tabular}
\label{tab:energy_estimate}
\end{table}

\textbf{Privacy and Ethical Considerations:} The proposed TinyNina framework relies exclusively on satellite-based multispectral imagery and aggregated environmental monitoring data. Sentinel-2 observations provide environmental measurements at spatial resolutions of 10-20 meters, which do not capture identifiable individuals or private activities. Consequently, the system does not involve personally identifiable information or street-level surveillance. Nevertheless, responsible deployment of satellite-based environmental monitoring systems requires transparency in model predictions and awareness of potential biases introduced by uneven spatial distribution of ground monitoring stations.

While TinyNina sacrifices general-purpose super-resolution performance to optimize NO\textsubscript{2} prediction accuracy, this is an intentional design choice. Our results demonstrate that in domain-specific applications such as environmental monitoring and intelligent transportation systems, task-aware design can outperform both model scale and traditional perceptual benchmarks. Importantly, TinyNina’s edge-ready design makes it suitable for integration into smart mobility infrastructures, including real-time deployment in connected vehicles for adaptive eco-routing, roadside IoT stations for emission-zone enforcement, and urban ITS control centers for traffic-light optimization. Future work may explore hybrid models that combine TinyNina’s efficiency with broader adaptability to other pollutants and remote sensing tasks, further strengthening its role in sustainable transportation and climate action strategies.

\section{Conclusion}
\label{sec:conclusion}

This study presents TinyNina, an ultra-lightweight super-resolution framework that overcomes key challenges in satellite-based NO\textsubscript{2} monitoring by reducing computational costs, eliminating reliance on external datasets, and prioritizing task-specific accuracy. Achieving a 7.4~$\mu g/m^3$ MAE with 95\% fewer parameters and 47$\times$ faster inference, TinyNina proves both efficient and scalable for real-time edge deployment. 

Beyond technical performance, TinyNina enables practical integration into sustainable urban planning, transportation emissions monitoring, and intelligent mobility infrastructures. Its deployment potential in connected vehicles, roadside IoT, and ITS control centers underscores its role in greener cities and climate-resilient policy. Overall, TinyNina demonstrates how efficient edge-AI models can bridge the gap between algorithmic innovation and sustainable societal impact.


\section*{Acknowledgment}
This research was funded by the Research Ireland Centre for Research Training in Digitally-Enhanced Reality (d-real) under Grant No. 18/CRT/6224. This research was conducted with the financial support of Science Foundation Ireland under Grant Agreement No.\ 13/RC/2106\_P2 at the ADAPT SFI Research Centre at University College Dublin. ADAPT, the SFI Research Centre for AI-Driven Digital Content Technology, is funded by Science Foundation Ireland through the SFI Research Centres Programme.

\bibliographystyle{IEEEtran.bst}
\bibliography{./ref/longforms,./ref/references}

\EOD

\end{document}